\documentclass{article}

\PassOptionsToPackage{numbers}{natbib}

\usepackage[final]{neurips_2018}

\usepackage[utf8]{inputenc} 
\usepackage[T1]{fontenc}    
\usepackage{hyperref}       
\usepackage{url}            
\usepackage{booktabs}       
\usepackage{amsfonts}       
\usepackage{nicefrac}       
\usepackage{microtype}      
\usepackage{epsfig}
\usepackage{graphicx}
\usepackage{amsmath}
\usepackage{amssymb}
\usepackage{subcaption}
\usepackage{float}
\usepackage{multirow}
\usepackage{comment}
\usepackage{wrapfig}

\usepackage[usenames, dvipsnames]{color}

\title{Dialog-based Interactive Image Retrieval}

\author{
  Xiaoxiao Guo$^{\dagger}$ \thanks{$^{\dagger}$These two authors contributed equally to this work. } \\
  IBM Research AI\\
  \texttt{xiaoxiao.guo@ibm.com}
  \And
  Hui Wu$^{\dagger}$ \\
  IBM Research AI\\
  \texttt{wuhu@us.ibm.com}
  \And
  Yu Cheng \\
  IBM Research AI\\
  \texttt{chengyu@us.ibm.com}
  \And
  Steven Rennie \\
  Fusemachines Inc.  \\
  \texttt{srennie@gmail.com}
  \And
 Gerald Tesauro \\
  IBM Research AI\\
  \texttt{gtesauro@us.ibm.com}
  \And
  Rogerio Schmidt Feris \\
  IBM Research AI\\
  \texttt{rsferis@us.ibm.com}
}

\renewcommand\footnotemark{}
\begin{document}

\maketitle

\begin{abstract}

Existing methods for interactive image retrieval have demonstrated the merit of integrating
user feedback, improving retrieval results. However, most current systems
rely on restricted forms of user feedback, such as binary relevance responses, or
feedback based on a fixed set of relative attributes, which limits their impact. In this
paper, we introduce a new approach to interactive image search that enables users
to provide feedback via natural language, allowing for more natural and effective
interaction. We formulate the task of dialog-based interactive image retrieval as a
reinforcement learning problem, and reward the dialog system for improving the
rank of the target image during each dialog turn. To mitigate the cumbersome and
costly process of collecting human-machine conversations as the dialog system
learns, we train our system with a user simulator, which is itself trained to describe
the differences between target and candidate images. The efficacy of our approach
is demonstrated in a footwear retrieval application. Experiments on
both simulated and real-world data show that 1) our proposed learning framework
achieves better accuracy than other supervised and reinforcement learning baselines
and 2) user feedback based on natural language rather than pre-specified
attributes leads to more effective retrieval results, and a more natural and expressive
communication interface.
 
\end{abstract}

\section{Introduction}

The volume of searchable visual media has grown tremendously in recent years, and has intensified the need for retrieval systems that can more effectively identify relevant information, with applications in domains such as e-commerce~\cite{huang2015cross,liuLQWTcvpr16DeepFashion},  surveillance~\cite{vaquero2009attribute,shi2015transferring}, and Internet search~\cite{gordo2016deep,jegou2012aggregating}. Despite significant progress made with deep learning based methods~\cite{wang2014learning,gordo2017end}, achieving high performance in such retrieval systems remains a challenge, due to the well-known semantic gap between feature representations and high-level semantic concepts, as well as the difficulty of fully understanding the user's search intent. 

\begin{figure}[t]
\begin{center}
\includegraphics[width=1\linewidth]{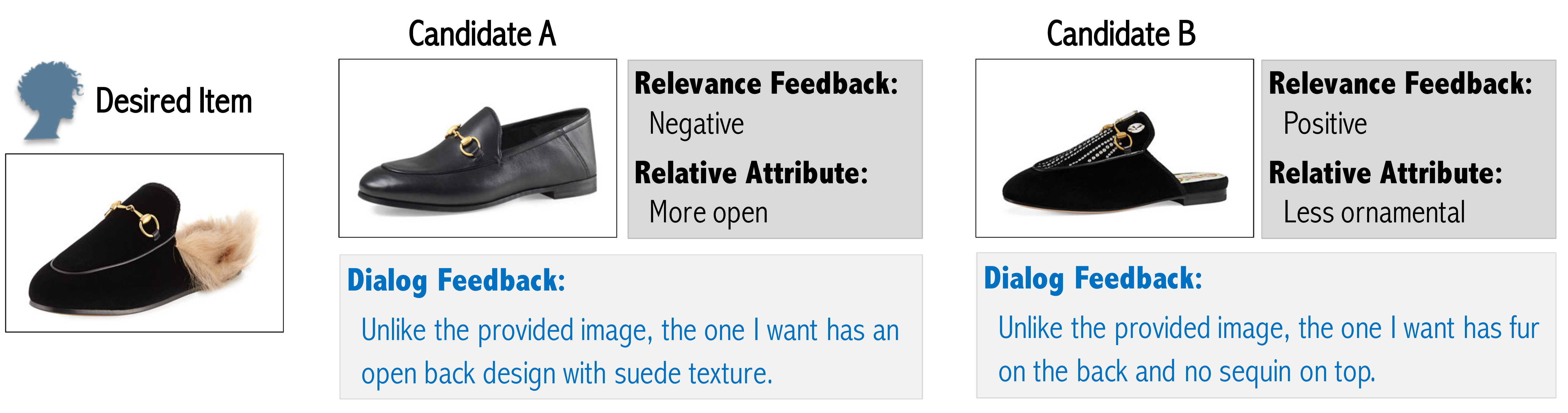}
\end{center}
\caption{In the context of interactive image retrieval, the agent incorporates the user's 
feedback to iteratively refine retrieval results. Unlike existing work which 
are based on relevance feedback or relative attribute feedback, our approach 
allows the user to provide feedback in natural language. 
\vspace{-0.05em}
}
\label{fig:teaser}
\end{figure}

A typical approach to improve search efficacy is to 
allow the user a constrained set of possible interactions with the system~\cite{zhou2003relevance,thomee2012interactive}. 
In particular, the user provides iterative feedback about retrieved 
objects, so that the system can refine the results, allowing the user and system 
to engage in a ``conversation'' to resolve what the user wants to retrieve.
For example, as shown in Figure~\ref{fig:teaser}, feedback about relevance~\cite{rui1998relevance} allows users to indicate which images are ``similar'' or ``dissimilar'' to the 
desired image, and relative attribute feedback~\cite{kovashka2012} allows
the comparison of the desired image with candidate images based on a fixed 
set of attributes. While these feedback paradigms are effective, the restrictions 
on the specific form of user interaction largely constrain the information that a 
user can convey to benefit the retrieval process.

In this work, we propose a new approach to interactive visual content retrieval by 
introducing a {\em novel form of user feedback based on natural language}. 
This enables users to directly express, in natural language, 
the most prominent conceptual differences between the preferred search object and 
the already retrieved content, which permits a more natural human-computer interaction. 
We formulate the task as a reinforcement learning (RL) problem, where the system 
directly optimizes the rank of the target object, which is a non-differentiable objective. 

We apply this RL based interactive retrieval framework to the task of image retrieval,
which we call {\em dialog-based interactive image retrieval} to emphasize its capability in aggregating history information compared to existing single turn approaches~\cite{tellex2009towards,barbu2013saying,li2017person,hu2016natural}.  In particular, a 
novel end-to-end dialog manager architecture is proposed, which takes natural language 
responses as user input, and delivers retrieved images as output. 
To mitigate the cumbersome and costly process of collecting and annotating 
human-machine dialogs as the system learns, 
we utilize a model-based RL approach by training a user simulator based on a corpus of human-written relative descriptions. Specifically, to emulate a single dialog turn, where the user provides feedback regarding a candidate image relative to what the user has in mind, the user simulator generates a {\em relative caption} describing the differences between the candidate image and the user's desired image.\footnote{In this work, the user simulator is trained on single-turn data and does not consider the dialog history. This reduces the sequence of responses to a ``bag'' of responses and implies that all sequences of a given set of actions (candidate images) are equivalent. Nevertheless, while the set of candidate images that maximize future reward (target image rank) are a set, selecting the image for the next turn naturally hinges on all previous feedback from the user. Therefore, the entire set of candidate
images can be efficiently constructed sequentially. }
Whereas there is a lot of prior work in image captioning \cite{kulkarni2011baby,vinyals2015show,rennie2016self}, 
we 
explore the problem of {\em relative image captioning}, a general approach to more expressive and natural communication of relative preferences to machines, and 
to use it as part of a user simulator to train a dialog system. 

The efficacy of our approach is evaluated in a real-world application
scenario of interactive footwear retrieval. Experimental results with both 
real and simulated users show that the proposed reinforcement learning
framework achieves better retrieval performance than existing techniques. 
Particularly, we observe that feedback based on natural language is 
more effective than feedback based on pre-defined relative attributes by a large
margin. Furthermore, the proposed RL training framework of directly optimizing 
the rank of the target image shows promising results and outperforms the supervised learning approach which is based on the triplet loss objective. 
The main contributions of this work are as follows:
\begin{itemize}
\item A new vision/NLP task and machine learning problem setting for interactive visual content search, where the dialog agent learns to interact with a human user over the course of several dialog turns, and the user gives feedback in natural language. 
\item A novel end-to-end deep dialog manager architecture, which addresses the above problem setting in the context of image retrieval. The network is trained based on an efficient policy optimization strategy,
employing triplet loss and model-based policy improvement~\cite{sutton1998reinforcement}.
\item The introduction of a 
computer vision task, \emph{relative image captioning}, 
where the generated captions describe the salient visual differences between two images, which is distinct from and complementary to context-aware \emph{discriminative image captioning}, where the absolute attributes of one image that discriminate it from another are described ~\cite{vedantam2017}.
\item The contribution of a new dataset, which supports further research on the task of relative image captioning. \footnote{The project website is at: \url{www.spacewu.com/posts/fashion-retrieval/}}
\end{itemize}

\section{Related Work}

\paragraph{Interactive Image Retrieval.}  Methods for improving image search results based on user interaction have been studied for more than 20 years~\cite{flickner1995query,thomee2012interactive,rui1999image}. Relevance feedback is perhaps the most popular approach, with user input specified either as binary feedback (``relevant'' or ``irrelevant'')~\cite{rui1998relevance} or based on multiple relevance levels~\cite{wu2004willhunter}. More recently, relative attributes (e.g., ``more formal than these,'' ``shinier than these'') have been exploited as a richer form of feedback for interactive image retrieval \cite{kovashka2012,kovashka2017attributes,kovashka2013attribute,parikh2011relative,yu2017fine}. 
All these methods rely on a fixed, pre-defined set of attributes, whereas our method relies on feedback based on natural language, providing more flexible and more precise descriptions of the items to be searched. Further, our approach offers an end-to-end training mechanism which facilitates deployment of the system in other domains, without requiring the explicit effort of building a new vocabulary of attributes.

\paragraph{Image Retrieval with Natural Language Queries.}
Significant progress has been recently made on methods that lie in the intersection of computer vision and natural language processing, such as image captioning~\cite{vinyals2015show,rennie2016self}, visual question-answering~\cite{antol2015vqa,tapaswi2016movieqa}, visual-semantic embeddings~\cite{frome2013devise,wang2016learning},  and grounding phrases in image regions~\cite{rohrbach2016grounding,plummer2015flickr30k}. In particular, our work is related to image or video retrieval methods based on natural language queries~\cite{tellex2009towards,barbu2013saying,li2017person,hu2016natural}. 
These methods, however, retrieve images and videos in a single turn, whereas our proposed approach aggregates history information from dialog-based feedback and iteratively provides more refined results.

\paragraph{Visual Dialog.} Building conversational agents that can hold meaningful dialogs with humans has been a long-standing goal of Artificial Intelligence.
Early systems were generally designed based on rule-based and slot-filling techniques \cite{williams2007partially}, whereas modern approaches have focused on end-to-end training, 
leveraging the success of encoder-decoder architectures and sequence-to-sequence learning \cite{serban2016building,bordes2016learning,guo2016learning}.
Our work falls into the class of visually-grounded dialog systems \cite{das2016visual,de2016guesswhat,seo2017visual,das2017learning,strub2017end}. Das et al \cite{das2016visual} proposed the task of visual dialog, where
the system has to answer questions about images based on a previous dialog history. 
De Vries et al. \cite{de2016guesswhat} introduced the {\em GuessWhat} game, where a series of questions are asked to pinpoint a specific object in an image, with restricted answers 
consisting of yes/no/NA.
The {\em image guessing} game \cite{das2017learning} 
demonstrated emergence of grounded language and communication
among visual dialog agents with no human supervision, using RL to train the agents in a goal-driven dialog setting.
However, these dialogs are purely text-based for both the questioner and answerer agents, whereas we address the interactive image retrieval problem,
with an agent presenting images to the user to seek feedback in natural language.


\section{Method}

Our framework
, which we refer to as the {\em dialog manager}, 
considers a user interacting with a retrieval agent via iterative dialog turns. 
At the $t$-th dialog turn, the dialog manager presents a candidate image $a_{t}$ selected from a retrieval database $ \mathcal{I} = \{I_i\}_{i=0}^{N}$ to the user. The user then provides a feedback sentence $o_{t}$, describing the differences between the candidate image $a_{t}$ and the desired image. Based on the user feedback and the dialog history up to turn $t$, $\mathrm{H}_{t}= \{a_{1},o_{1},...,a_{t},o_{t} \}$, the dialog manager selects another candidate image $a_{t+1}$ from the database and presents it to the user. This process continues until the desired image is selected or the maximum number of dialog turns is reached. 
In practice, the dialog manager
could provide multiple images per turn to achieve better retrieval
performance. In this work, we focus on a simplified scenario with a single image 
per interaction. We
note that the same framework could be extended to
the multiple-image case by allowing the user to select one
image out of a list of candidate images to provide feedback on.

\begin{figure*}[t!]
\begin{center}
\includegraphics[width=\linewidth]{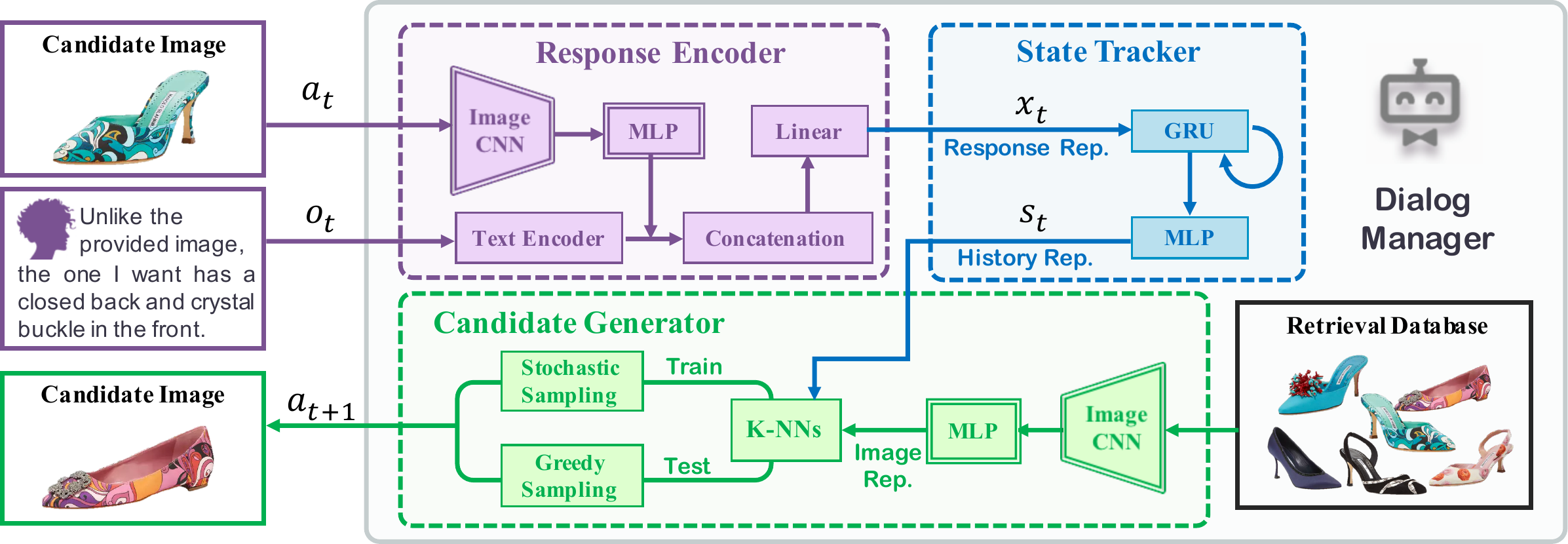}
\end{center}
\caption{The proposed end-to-end framework for dialog-based interactive image retrieval. }
\label{fig:framework}
\end{figure*}

\subsection{Dialog Manager: Model Architecture}
\label{sec:framework}
Our proposed dialog manager model consists of three main components: a {\em Response Encoder}, a {\em State Tracker}, and a {\em Candidate Generator}, as shown in Figure~\ref{fig:framework}. At the $t$-th dialog turn, the {\em Response Encoder} embeds a candidate image and the corresponding user feedback $\{a_t, o_t\}$ into a joint visual-semantic representation $x_t \in \mathbb{R}^D$. The {\em State Tracker} then aggregates this representation with the dialog history from previous turns, producing a new feature vector $s_t \in \mathbb{R}^D$ . The {\em Candidate Generator} uses the aggregated representation $s_t$ to select a new candidate image $a_{t+1}$ that is shown to the user. Below we provide details on the specific design of each of the three model components.

\paragraph{Response Encoder.} 
The goal of the Response Encoder is to embed the information from the $t$-th dialog turn 
$\{a_{t},o_{t}\}$ into a visual-semantic representation $x_t \in \mathbb{R}^D$. First, the candidate image is encoded using a deep convolutional neural network (CNN) followed by a linear transformation: $x_t^{\textrm{im}} = \textrm{ImgEnc}(a_t) \in \mathbb{R}^{D}$. The CNN architecture in our implementation is an ImageNet pre-trained ResNet-101~\cite{He2015} and its parameters are fixed. 
Words in the user feedback sentence are represented with one-hot vectors and then embedded with a linear projection followed by a CNN as in \cite{kim2014convolutional}: 
$x_t^{\textrm{txt}} = \textrm{TxtEnc}(o_t) \in \mathbb{R}^{D}$. Finally, the image feature vector and 
the sentence representation 
are concatenated and embedded through a linear transformation to obtain the final 
response representation at time $t$: 
$x_{t}=W(x^{\textrm{im} }_{t} \oplus x^{\textrm{txt}}_{t} )$, 
where $\oplus$ is the concatenation operator and $W \in \mathbb{R}^{D \times 2D}$ is the linear 
projection. The learnable parameters of the Response Encoder are denoted as $\theta_{r}$. 

\paragraph{State Tracker.} The State Tracker is based on a gated recurrent unit (GRU), 
which receives as input the response representation $x_t$, combines it with
the history representation of previous dialog turns, and outputs the 
aggregated feature vector $s_t$.
The forward dynamics of the State Tracker is written as: $g_{t}, h_{t}= \textrm{GRU} (x_{t}, h_{t-1})$, $s_{t} = W^{s} g_{t}$, 
where $h_{t-1} \in \mathbb{R}^{D}$ and $g_{t} \in \mathbb{R}^{D}$ are the hidden 
state and the output of the GRU respectively, $h_{t}$ is the updated hidden state,
$W^{s} \in \mathbb{R}^{D \times D}$ is a linear projection and $s_t\in \mathbb{R}^{D}$ is the history representation updated with the 
information from the current dialog turn. 
The learnable parameters of the State Tracker (GRU model) are denoted as $\theta_{s}$. 
This memory-based design of the State Tracker allows our model to sequentially aggregate the information from user feedback to localize the candidate image
to be retrieved.

\paragraph{Candidate Generator.} 
Given the feature representation of all images from the retrieval database, 
$\{x_i^{\textrm{im}}\}_{i=0}^{N}$, where $x_i^{\textrm{im}} = \textrm{ImgEnc}(I_i)$, 
we can compute a sampling probability based on the distances between
the history representation $s_{t}$ to each image feature, $x_i^{\textrm{im}}$.
Specifically, the sampling probability is modeled 
using a softmax distribution over the top-$K$ nearest neighbors
of $s_t$: $\mathbf{\pi}(j) = e^{-{d_j}} / \sum_{k=1}^{K}e^{-{d_k}}, j=1,2,...,K$, where $d_k$ is the $L2$ distance of $s_t$ to its $k$-th nearest neighbor in 
$\{x_i^{\textrm{im}}\}_{i=0}^{N}$. Given the sampling distribution, 
two approaches can be taken to sample the candidate image, denoted as $a_{t+1} = I_{j^{\prime}}$: 
(1) stochastic approach (used at training time), where $j^{\prime} \sim \mathbf{\pi}$, and (2) greedy
approach (used at inference time), where $j^{\prime} = \arg\max_{j}(\pi_{j})$. 
Combining the three components of the model architecture, the overall learnable
parameters of the dialog manager model is $\theta=\{\theta_{r}, \theta_{s}\}$. 
Next, we explain how the network is trained end-to-end.

\begin{figure*}[ht!]
\centering
    \begin{subfigure}[t]{0.45\textwidth}
        \centering
        \includegraphics[width=1\textwidth]{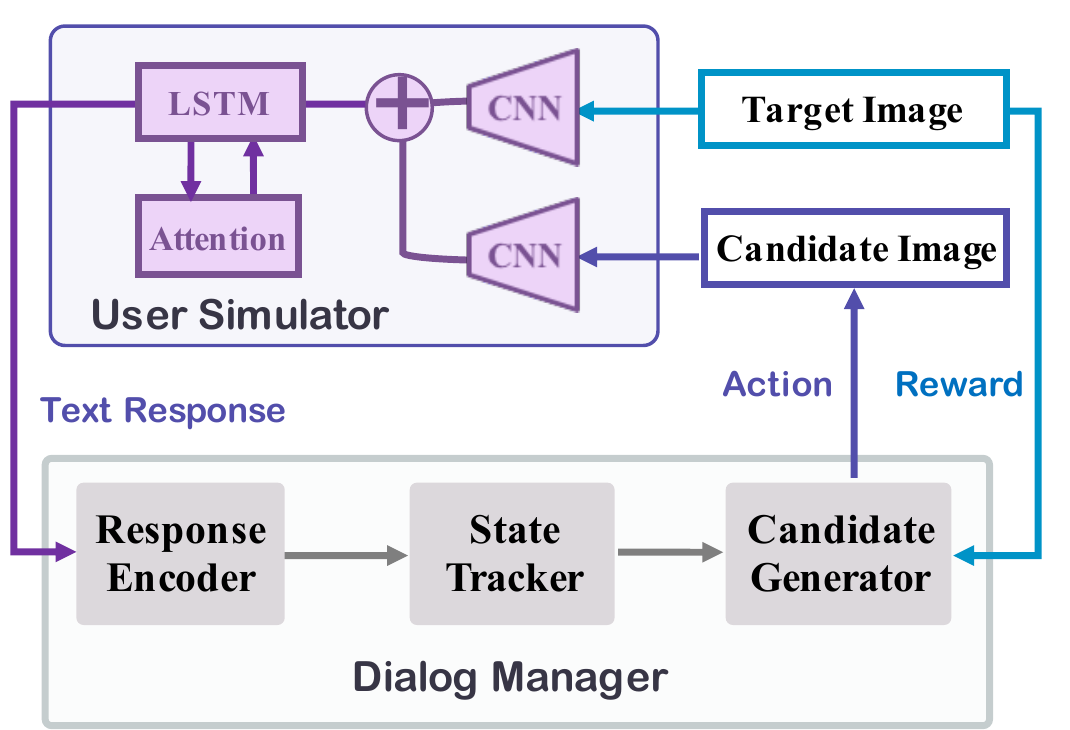}
        \caption{}
    \end{subfigure}%
    ~ 
    \begin{subfigure}[t]{0.55\textwidth}
        \centering
        \includegraphics[width=1\textwidth]{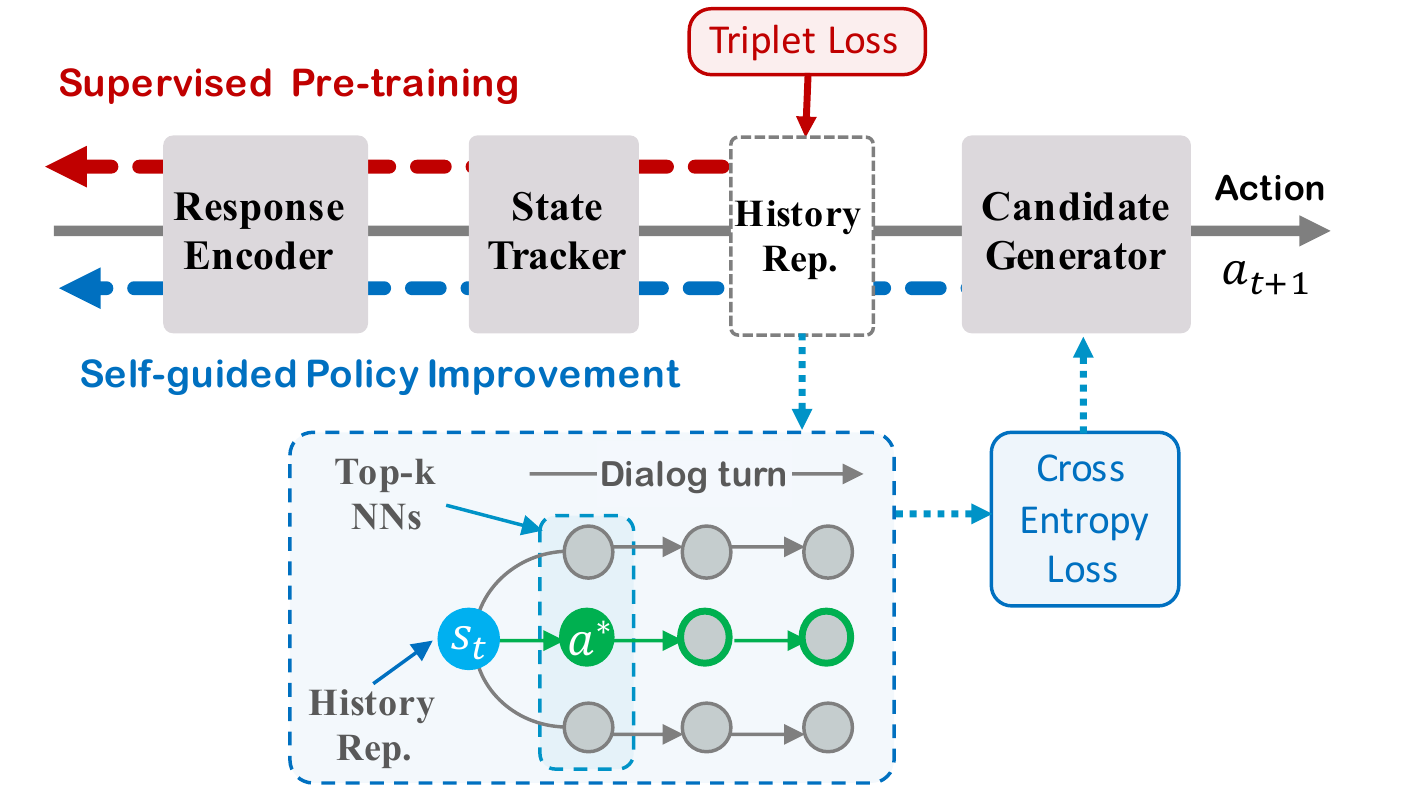}
        \caption{}
    \end{subfigure}
    \caption{The learning framework: (a) The user simulator enables efficient 
    exploration of the retrieval dialogs (Section~\ref{sec:simulator}); 
    and (b) the policy network is learned using a combination of supervised 
    pre-training and model-based policy improvement (Section~\ref{sec:optimization}).}
    \label{fig:pipeline}
\end{figure*}

\subsection{Training the Dialog Manager}
\label{sec:drl}

Directly optimizing the ranking percentile metric in a supervised learning scheme
is challenging since it is a non-differentiable function~\cite{li2016multiple,wang2018lambdaloss}. 
Instead, we model the ranking percentile as the environment reward received by the agent and
frame the learning process in a reinforcement learning setting with the goal of 
maximizing the expected sum of discounted rewards:
$\max_{\pi} \mathcal{U^{\pi}} = \mathbb{E} \big[ \sum_{t=1}^{T} \gamma^{t-1} r_{t} | \pi_\theta \big]$, where $r_{t}\in \mathbb{R}$ is the reward representing the ranking percentile 
of the target image at the $t$-th interaction, $\gamma$ is a discount factor determining 
the trade-off between short-term and long-term rewards, 
$T$ is the maximum number 
of dialog turns, 
and $\pi_\theta$ is the policy determined by network parameters $\theta$.\footnote{Strictly speaking, the optimal policy depends on the number of remaining dialog turns. We simplify the policy to be a function independent of dialog turn numbers.} 

Training an RL model for this problem requires extensive exploration 
of the action space, which is only feasible if a large amount of training data is available. 
However, collecting and annotating human-machine dialog data for our task is expensive. 
This problem is exacerbated in the situation of natural language based user feedback, 
which incurs an even larger exploration space as compared to approaches based on a fixed set of attributes. 
In text-based dialog systems, it is common to circumvent this 
issue by relying on user simulators \cite{li2016user}. We adopt a similar strategy, 
where a user simulator, trained on human-written relative descriptions, substitutes the 
role of a real user in training the dialog manager (illustrated in Figure~\ref{fig:pipeline}a). 
Below we further describe our user simulator, as well as the reinforcement learning techniques that we used to optimize our dialog manager.

\subsubsection{User Simulator based on Relative Captioning}
\label{sec:simulator}
Here we propose the use of a {\em relative captioner} to simulate the user. 
It acts as a surrogate for real human 
users by automatically generating sentences that can describe the 
prominent visual differences between any pair of target and candidate images. 
We note that at each turn, our user simulator generates feedback independent of previous user feedback, and previously retrieved images. While more sophisticated models that consider the dialog history could potentially be beneficial, training such systems well may require orders of magnitude more annotated data. In addition, back-referencing in dialogs can inherently be ambiguous and complex to resolve, even for humans. Based on these considerations, we decided to first investigate the use of a single-turn simulator. 
While a few related tasks have been studied previously, such as context-aware image captioning~\cite{vedantam2017} and referring expression generation~\cite{yu2016}, to the best of our knowledge, 
there is no existing 
dataset  directly supporting this task, so we introduce a new dataset as described in Section~\ref{sec:dataset}.

We experimented with several different ways of combining the visual features of the target and retrieved images. We include a comprehensive study of different models for the user simulator in Appendix~\ref{sec:app_relative} and show that the relative captioner 
based user model serves as a reasonable proxy for real users.
Specifically, we used feature concatenation to fuse the image features of 
the target and the reference image pair and applied the model of {\em Show, 
Attend, and Tell}~\cite{icml2015_xuc15} to generate the relative captions
using a long short-term memory network (LSTM). 
For image feature extraction, we adopted the architecture of ResNet101~\cite{He2015}
pre-trained on ImageNet; and to better capture the localized visual differences, we added a visual attention mechanism; the loss function
of the relative captioner is the sum of the negative log likelihood of the 
correct words~\cite{icml2015_xuc15}. 
 
\subsubsection{Policy Learning} 

\paragraph{Supervised Pre-training.} 
When the network parameters are randomly initialized at the beginning, 
the history representations $s_{t}$ are nearly random. 
To facilitate efficient exploration during RL training, we first 
pre-train the policy using a supervised learning objective.
While maximum likelihood-based pre-training is more common, here we pre-train using the more discriminative triplet loss objective:

\begin{equation}
\begin{aligned}
\vspace{-2em}
\mathcal{L}^{\textrm{sup}} &= \mathbb{E} \Big[ \sum_{t=1}^{T} \max(0, \|s_{t} 
- x^{+}\|_2 - \|s_{t} - x^{-}\|_2 + \text{m})  \Big]
\end{aligned}
\end{equation}
where $x^{+}$ and $x^{-}$ are the image features of the target image and a random 
image sampled from the retrieval database respectively, $\text{m}$ is a constant for the margin and $\|.\|_2$ denotes $L2$-norm. 
Intuitively, by ensuring the proximity of the target image and the images 
returned by the system, the rank of the target image can be improved 
without costly policy search from random initialization. However, 
entirely relying on this supervised learning objective deviates from our 
main learning objective, since the triplet loss objective does not 
jointly optimize the set of candidate images to maximize expected future 
reward.~\footnote{More explanation on the difference between the two objectives
is provided in Appendix~\ref{sec:qualitative}.}

\paragraph{Model-Based Policy Improvement.}  
\label{sec:optimization}
Given the known dynamics of the environment (in our case, the user simulator), it is often
advantageous to leverage its behavior for policy improvement. Here we adapt the policy improvement~\cite{sutton1998reinforcement} to our model-based policy learning.
Given the current policy $\pi$ and the user simulator, the value
of taking an action $a_{t}$ using test-time configuration can be efficiently 
computed by look-ahead policy value estimation 
$Q^{\pi}(h_{t}, a_{t}) = \mathbb{E} \big[ \sum_{t'=t}^{T} \gamma^{t'-t} r_{t'} | \pi \big]$. Because our user simulator is essentially deterministic, one trajectory 
is sufficient to estimate one action value.
Therefore, an improved policy $\pi'$ can be derived from the current policy $\pi$ 
by selecting the best action given the value of the current policy, $\pi'(h_{t}) \equiv a^{*}_{t} = \arg \max_{a} Q^{\pi}(h_{t}, a)$. 
Specifically, following ~\cite{guo2014deep}, we minimize the cross entropy loss given 
the derived action, $a^{*}$,
\begin{equation}
\begin{aligned}
\mathcal{L}^{\textrm{imp}} = \mathbb{E} \Big[ - \sum_{t=1}^{T} \log\Big(\pi(a^{*}_{t}|h_{t})\Big)\Big]
\end{aligned}
\end{equation}
Compared to traditional policy gradient methods, the model-based policy improvement gradients have lower variance, and converge faster. In Section~\ref{sec:experiment}, we further demonstrated the effectiveness of model-based policy improvement by comparing it
with a recent policy gradient method. Figure~\ref{fig:pipeline}b illustrates our policy learning method as described above.

\nocite{li2014req}



\section{Dataset: Relative Captioning}
\label{sec:dataset}

Our user simulator aims 
to capture the rich and flexible language describing visual 
differences of any given image pair. The relative captioning dataset 
thus needs this property. We situated the data collection procedure
in a scenario of a shopping chatting session between a shopping assistant and a customer. The annotator was asked to take the role of the customer 
and provide a natural expression to inform the shopping assistant about the desired
product item. 
To promote more regular, specific, and relative user feedback, we provided a sentence prefix for the annotator to complete when composing their response to a retrieved
image. Otherwise the annotator response is completely free-form: no other constraints on the response were imposed.
We used Amazon Mechanical Turk 
to crowdsource the relative expressions. After manually removing erroneous annotations, 
we collected in total $10,751$ captions, with one caption per pair of images.

Interestingly, we observed that when the target image and the reference image are sufficiently different, users often directly describe the visual appearance of the target image, rather than using relative expressions (c.f. fourth example in Figure 7(b), Appendix A). This behavior mirrors the \emph{discriminative captioning} problem considered in \cite{vedantam2017}, where a method must take in two images and produce a caption that refers only to one of them. Relative and discriminative captioning are complementary, and in practice, both strategies are used, and so we augmented our dataset by pairing 3600 captions that were discriminative with additional dissimilar images. Our captioner and dialog-based interactive retriever are thus trained on both discriminative and relative captions, so as to be respectively more representative of and responsive to real users. Additional details about the dataset collection procedure and the analysis on dataset statistics are included in Appendix~\ref{sec:app_data} and Appendix~\ref{sec:dataset_analysis}.

\section{Experimental Results}
\label{sec:experiment}

In Section~\ref{sec:rlResult}, we assess the contribution of each component of our pipeline for policy learning. To evaluate the value of using free-form dialog feedback, 
we show experiments considering both simulated user feedback (Section~\ref{sec:dialogResult}) 
and real-world user feedback (Section~\ref{sec:humanResult}). 

All experiments were performed on the {\em Shoes} 
dataset~\cite{berg2010}, with the same training and testing data split
for all retrieval methods and for training the user simulator. 
$10,000$ database images were used during training, and $4,658$
images for testing. The retrieval models are tested by retrieving images on the testing set, starting from a randomly selected candidate image for the first dialog turn. 
Image retrieval performance is quantified by the average rank percentile 
of the image returned by the dialog manager on the test set.
For details on architectural configurations, parameter settings, 
baseline implementation, please refer to Appendix~\ref{sec:config}. 

\begin{wrapfigure}{r}{7cm}
\vspace*{-1.5em}
\includegraphics[width=7cm]{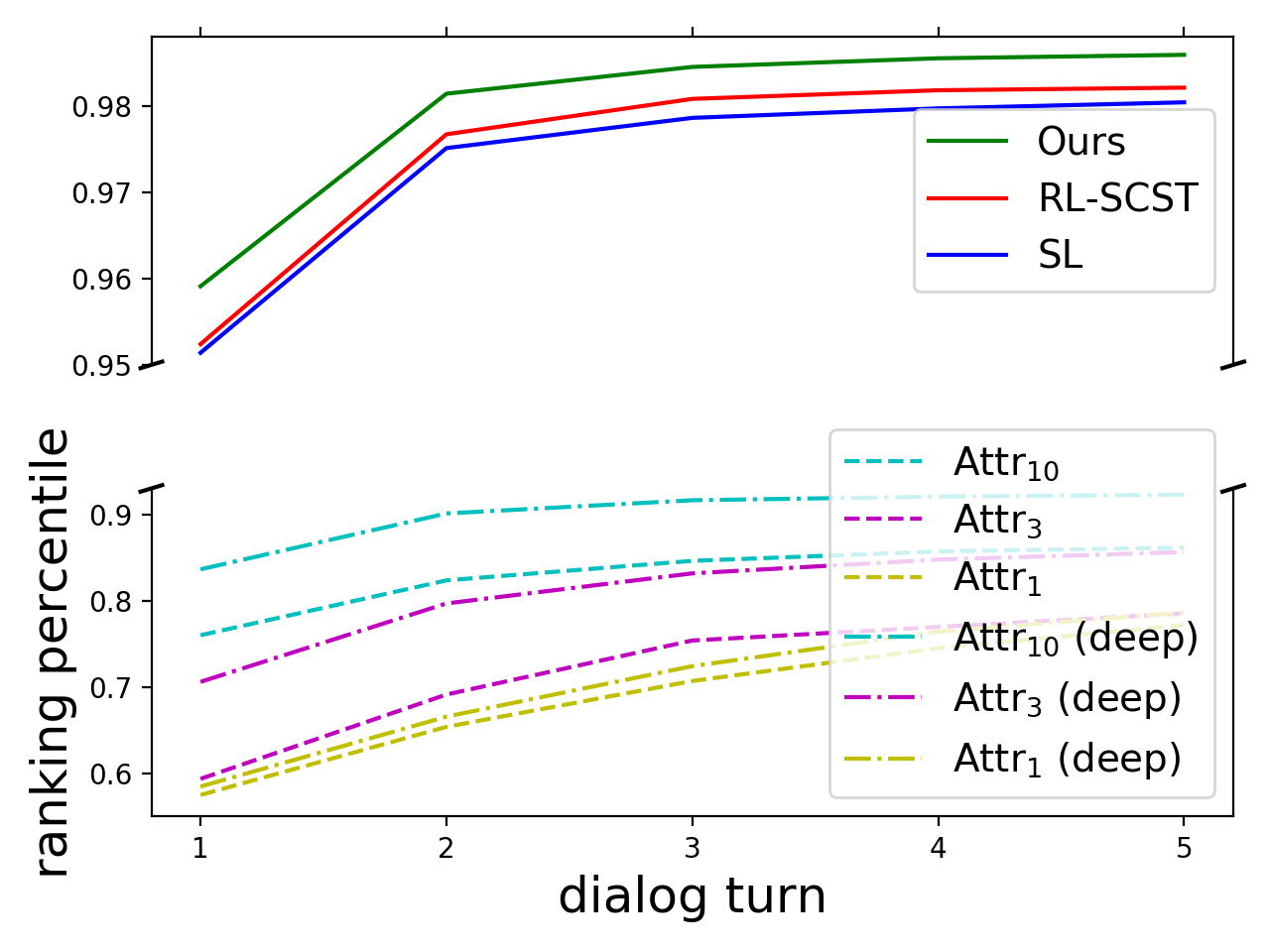}
\caption{
Quantitative comparison of our method and two baselines and
methods using feedback based on a pre-defined set of relative attributes. }
\label{fig:quan_result}
\vspace*{-1.5em}
\end{wrapfigure} 

\subsection{Analysis of the Learning Framework}
\label{sec:rlResult}
We use our proposed user simulator to generate data and provide
extensive quantitative analysis 
on the contribution
of each model component.

\begin{figure*}
\centering
\includegraphics[width=\textwidth]{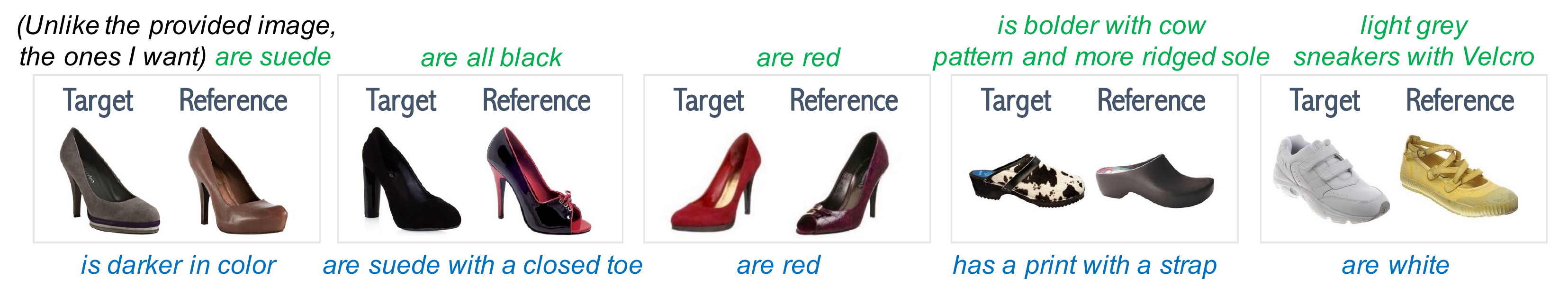} 
\caption{Examples of human provided (\textcolor{Green}{green}) and captioner generated relative descriptions 
(\textcolor{NavyBlue}{blue}). 
While generated relative captions don't resemble human annotations in 
most cases, they can nonetheless capture the main visual differences between the
target image and the reference image.}
\label{fig:captioner}
\end{figure*}

\paragraph{Results on Relative Captioner.} Figure~\ref{fig:captioner} provides examples of simulator generated 
feedback and the collected user annotations. An interesting 
observation is that even though the user simulator only occasionally 
generates descriptions that exactly match the human annotations 
(the third example in Figure~\ref{fig:captioner}),
it can still summarize the main visual differences
between the images, since inherently there are many ways 
to describe differences between two images. Qualitative 
examination of the generated relative expressions showed that the user 
simulator can approximate feedback of real users at a very low annotation cost 
(more analysis is included in Appendix~\ref{sec:app_relative}).

\paragraph{Policy Learning Results.} 
To investigate how retrieval performance is affected by each component 
of the dialog manager, we compare our approach, denoted as \textbf{Ours}, against 
two variants: (1) \textbf{SL}: supervised learning where the agent is trained only with triplet loss; 
(2) \textbf{RL-SCST}: policy learning using Self-Critical Sequence Training (SCST)~\cite{rennie2016self} after pre-training the network using the triplet loss objective. 
As shown in Figure~\ref{fig:quan_result} (solid lines), the average ranking percentile 
of the target image in all methods increases monotonically as the 
number of dialog turns increases. Both RL-based retrieval algorithms 
outperform the supervised pre-training, \textbf{SL}, which is 
expected since the triplet loss function does not directly optimize the retrieval
ranking objective. Finally, \textbf{Ours} achieves $98\%$ average ranking percentile 
with only two dialog turns and consistently outperforms \textbf{RL-SCST} 
across different dialog turns, which demonstrates the benefit of the model-based
policy improvement component. We have observed similar results on the attribute-based baselines. Each of the SL based attribute model underperforms its RL version by
 $\sim1\%$ in retrieval ranking percentile.

\subsection{Effectiveness of Natural Language Feedback}
\label{sec:dialogResult}

In this section, we empirically evaluate the effect of natural language feedback, compared
to pre-defined, relative attribute-based user feedback. 

{\noindent \bf Generating Attribute Feedback.} 
Each image in the 
dataset
maps to a $10$-D attribute vector, as described in~\cite{kovashka2012}.
We adopted a rule-based feedback generator which concatenates the 
respective attribute words with ``more'' or ``less'', depending on the relative attribute
values of a given image pair. For example, 
if the ``shiny'' value of the candidate image and the target image
are $0.9$ and $1.0$ respectively, then the rule-based feedback is 
``more shiny.''  Attributes are randomly sampled, similar to the relative attribute feedback generation in \cite{kovashka2012}. To simulate the scenario when users 
provide feedbacks using multiple attributes, individual attribute phrases are combined.  
We adopted original attribute values in~\cite{kovashka2012}, which
were predicted using hand-crafted image features, as well as attribute values predicted
using deep neural networks in~\cite{7494596}. 

{\noindent \bf Results.} 
We trained the dialog manager using both dialog-based feedback and attribute-based 
feedback (\textbf{Attr\textsubscript{n}} and \textbf{Attr\textsubscript{n} (deep)}), where the subscript number denotes
the number of attributes used in the rule-based feedback generator and \textbf{(deep)} denote baselines using deep learning based attribute estimates as in~\cite{7494596}. The empirical result
is summarized in Figure~\ref{fig:quan_result}, including relative attribute feedback using
one, three and ten attribute phrases. The three attribute case matches the average length of user feedback in free-form texts and the ten case uses all possible pre-defined attributes to provide feedback.   
Across different numbers of dialog turns, the natural language 
based agent produced significantly higher target image average 
ranking percentile than the attribute based
methods. 
The results suggest that feedback based on unrestricted natural language 
is more effective for retrieval than the predefined set of relative attributes used in \cite{kovashka2012}. This is expected as the vocabulary of relative attributes in \cite{kovashka2012} is limited.
Even though deep learning based attribute estimates improve the attribute-based baselines significantly, the performance gap between attribute based baseline and free form texts is still significant. We conjecture that the main reason underlying the  performance gap between attribute and free-form text based models is the effectively open domain for
attribute use, which is difficult to realize in a practical user interface without natural language. In fact, free-form dialog feedback obviates
constructing a reliable and comprehensive attribute taxonomy,
which in itself is a non-trivial task~\cite{maji2012}.

\subsection{User Study of Dialog-based Image Retrieval}
\label{sec:humanResult}
In this section, we demonstrate the practical use of our system 
with real users. We compare with an existing method, WhittleSearch~\cite{kovashka2012}, on the task of interactive footwear retrieval.  WhittleSearch represents images as feature 
vectors in a pre-defined $10$-D attribute space, and iteratively refines 
retrieval by incorporating the user's relative feedback on 
attributes to narrow down the search space of the target 
image. For each method, we collected $50$ five-turn dialogs; 
at each turn, one image is presented to the user to seek feedback. 
For WhittleSearch, the user can choose to use
any amount of attributes to provide relative feedback on during each
interaction. The resulting average ranking percentile of the dialog manager and 
WhittleSearch are $89.9\%$ and $70.3\%$ 
respectively. In addition to improved retrieval accuracy, users also 
reported that providing dialog-based feedback is more natural compared to selecting
the most relevant attributes from a pre-defined list.

\begin{figure}
\centering
\includegraphics[width=\linewidth]{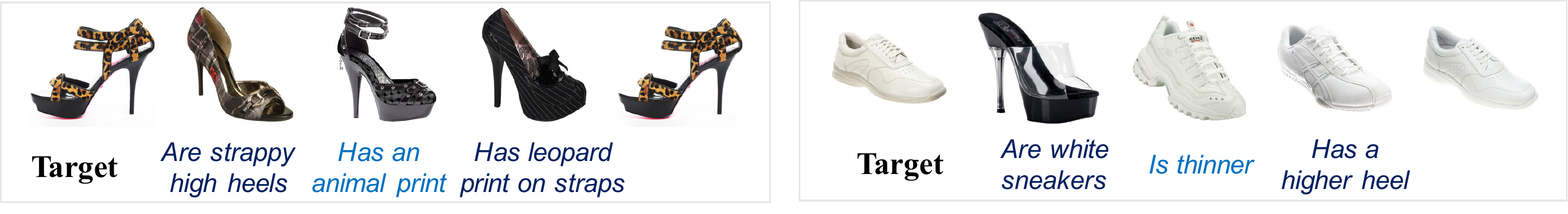} 
\caption{Examples of the user interacting with the proposed dialog-based 
image retrieval framework. }
\label{fig:userDialogs}
\vspace{-1em}
\end{figure}

Figure~\ref{fig:userDialogs} shows examples of retrieval dialogs from real users
(please refer to Appendix~\ref{sec:qualitative} for more results and discussions). 
We note that users often start the dialog with a coarse description of the main
visual features (color, category) of the target. As the dialog progresses, 
users give more specific feedback on fine-grained visual differences. 
The benefit of free-form dialog can be seen from the
flexible usage of rich attribute words (``leopard print on straps''),
as well as relative phrases (``thinner'', ``higher heel'').
Overall, these results show that the proposed framework for the
dialog manager exhibits promising behavior on generalizing to real-world applications. 
 
\section{Conclusions}
This paper introduced a novel and practical task residing at the intersection 
of computer vision and language understanding: 
dialog-based
interactive image retrieval.
Ultimately, techniques that are successful on such tasks will form the basis 
for the high fidelity,  multi-modal, intelligent 
conversational systems of the future, and thus represent important milestones in this quest.  
We demonstrated the value of 
the proposed learning architecture on the application
of interactive fashion footwear retrieval.
Our approach, 
enabling users to provide natural language feedback,
significantly outperforms
traditional methods relying on a pre-defined vocabulary of relative attributes, while offering more natural communication.
As future work, we plan to leverage side information, such as textual descriptions associated with images of product items,
and to develop user models that are
conditioned on dialog histories, enabling more realistic interactions.
We are also optimistic that our approach for image retrieval can be extended to other media types such as audio, video, and e-books, given the 
performance of deep learning on tasks such as speech recognition, machine translation, and activity recognition.

\textbf{Acknowledgement.}
We would like to give special thanks to Professor Kristen Grauman for helpful discussions.

\begingroup
   	\small
    \bibliographystyle{unsrt}
	\bibliography{nips2018}
\endgroup


\clearpage 
\newpage
\begin{center}
{\bf {\Large Supplemental Material: Dialog-based Interactive Image Retrieval \\} }
\end{center}
\appendix
\section{Data Collection}
\label{sec:app_data}

In the following, we explain the details on how we collected the relative 
captioning dataset for training the user simulator and provide insights 
on the dataset properties. Unlike existing datasets which aim to capture 
the visual differences purely using ``more" or ``less" relations on visual attributes~\cite{kovashka2012}, we want to collect data which 
captures comparative visual differences that are 
hard to describe merely using a pre-defined set of attributes. 
As shown in Figure~\ref{fig:amtInterface}, we designed the data collection 
interface in the context of fashion footwear retrieval, where 
a conversational shopping assistant interacts with a customer and whose goal
is to efficiently retrieve and present the product that matches 
the user's mental image of the desired item. 

\begin{figure*}[h]
\centering
	\begin{subfigure}[t]{0.4\textwidth}
        \centering
        \includegraphics[width=\linewidth]{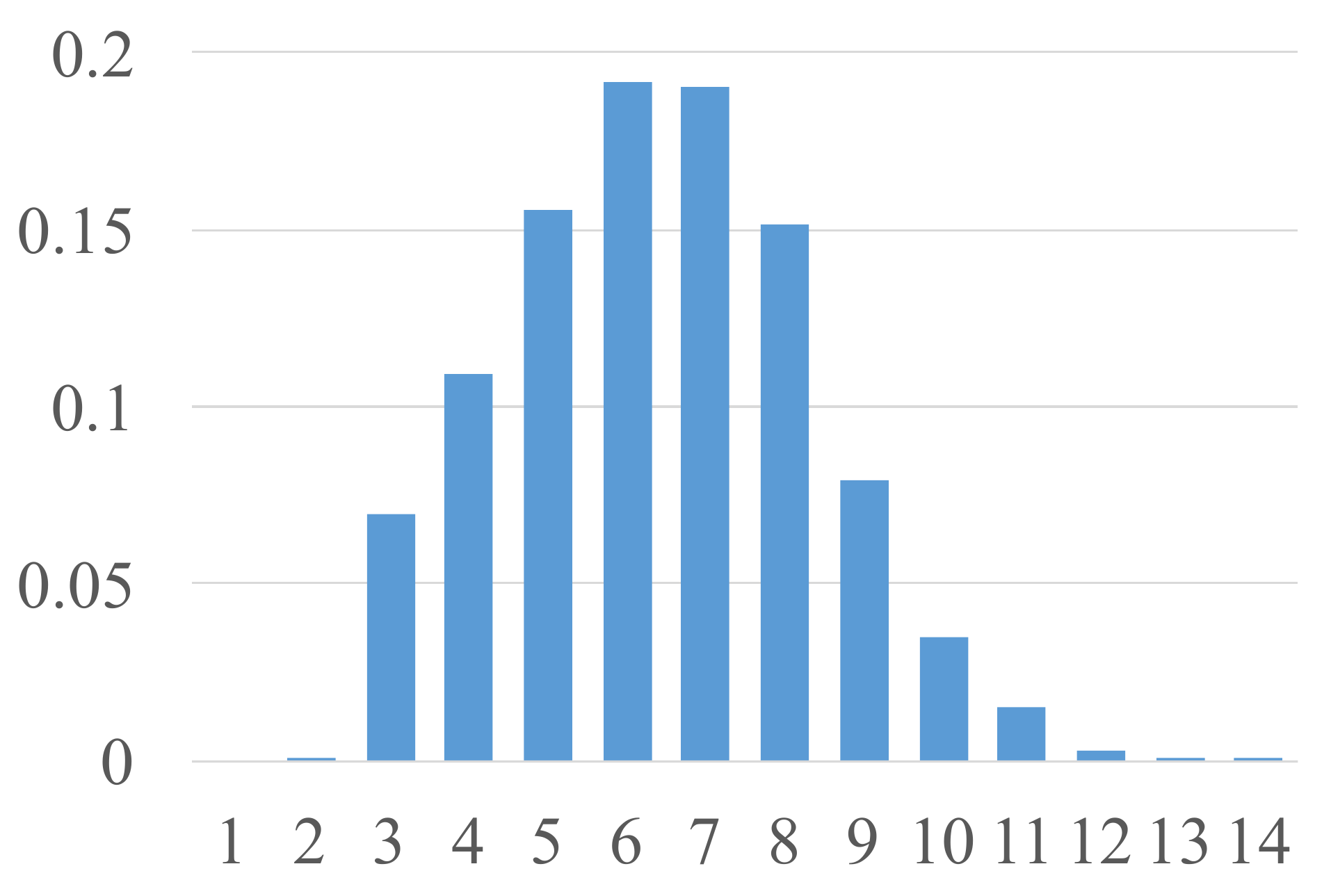}
        \caption{}
    \end{subfigure}
    \begin{subfigure}[t]{0.55\textwidth}
        \centering
        \includegraphics[width=\linewidth]{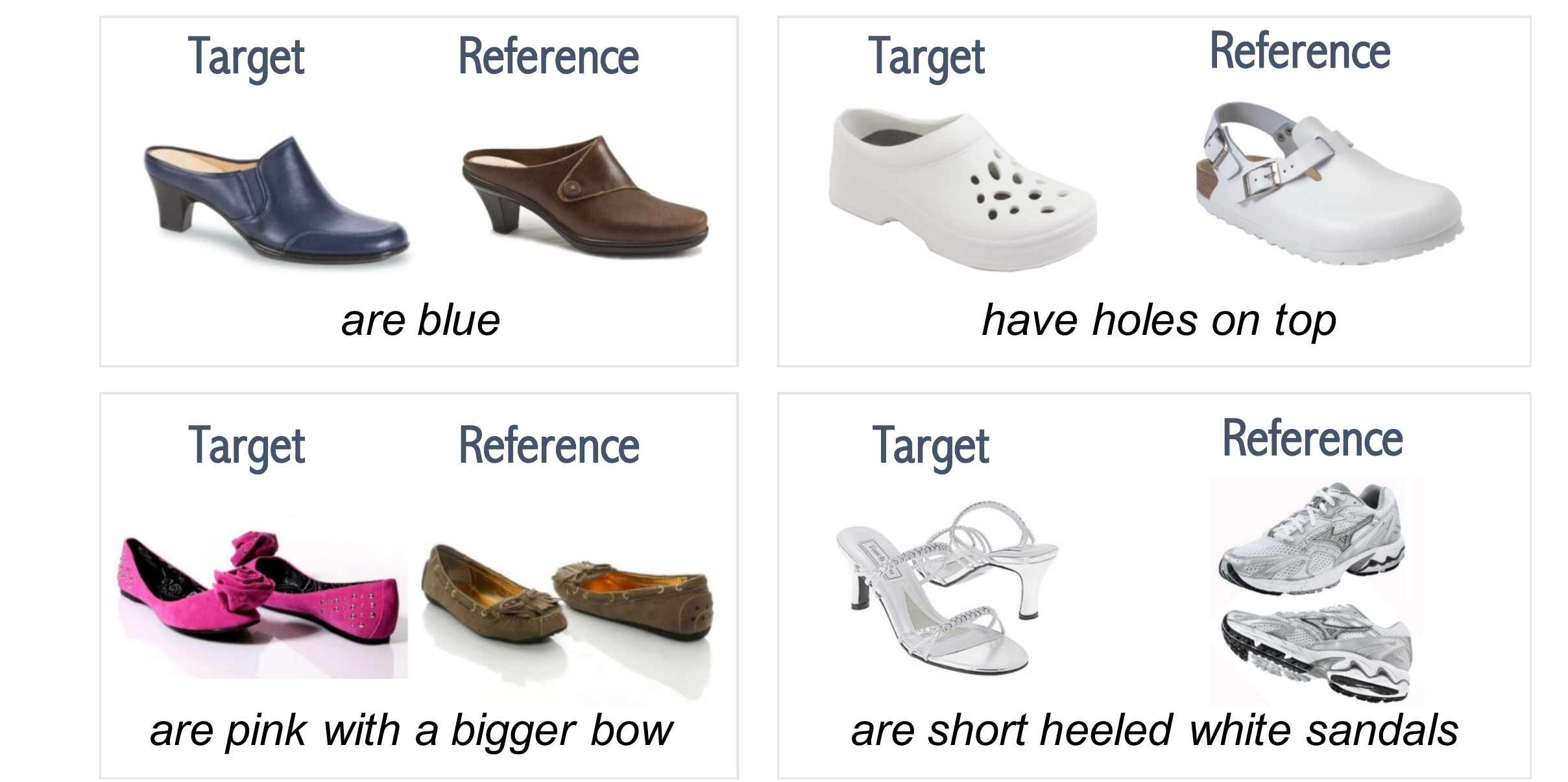}
        \caption{}
    \end{subfigure}%
    \caption{Length distribution of the relative captioning dataset (a),
    and examples of relative captions collected in the dataset (b). 
    The leading phrase ``\emph{Unlike the provided image, the ones I want}" 
    is omitted for brevity.}
    \label{fig:relativeExample}
\end{figure*}
  
\textbf{Collecting Relative Expressions.} 
The desired annotation for relative captioning should be free-form and
introduce minimum constraints on how a user might construct the feedback
sentence. On the other hand, we want the collected feedback to be concise and 
relevant for retrieval and avoid casual and non-informative 
phrases (such as ``\emph{thank you}", ``\emph{oh, well}"). 
Bearing the two goals in mind, we designed a data collection interface as
shown in Figure~\ref{fig:amtInterface}, which provided the beginning phrase 
of the user's response (``\emph{Unlike the provided ...}'') and the annotators only 
needed to complete the sentence by giving an informative relative expression. 
This way, we can achieve a balance between sufficient lexical flexibility and 
avoiding irrelevant and casual phrases. After manual data cleaning, we are left
with $10,751$ relative expressions with one annotation per image pair. 

\begin{figure}
\begin{center}
\includegraphics[width=10cm]{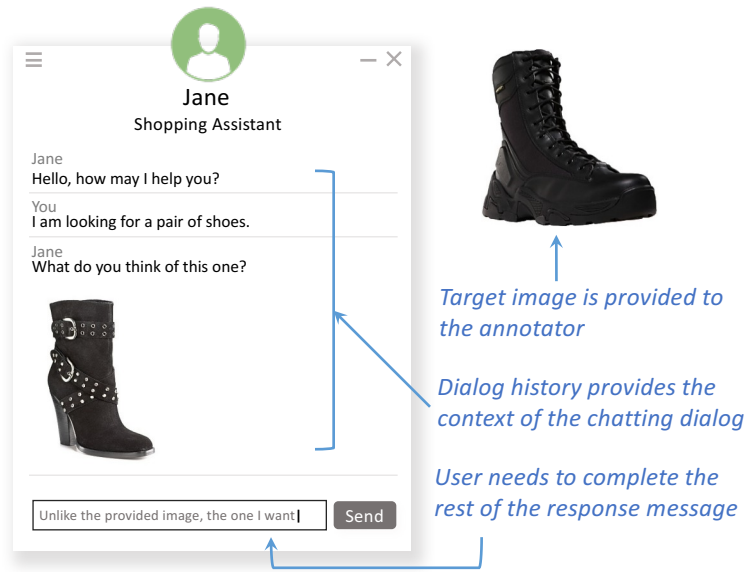}
\caption{AMT annotation interface. Annotators need to assume the role of the customer and
 complete the rest of the response message. The collected captions are concise,
and only contain phrases that are useful for image retrieval.}
\label{fig:amtInterface}
\end{center}
\end{figure} 

\textbf{Augmenting Dataset with Single-Image Captions.} 
During our data collection procedure for relative expressions, 
we observed that when the target image and the reference image are 
visually distinct (fourth example in Figure~\ref{fig:relativeExample}(b)), 
users often only implicitly use the reference image by directly describing
the visual appearance of the target image. 
Inspired by this, we asked annotators to give 
direct descriptions on $3600$ images without the use of reference images. 
We then paired each image in this set with 
multiple visually distinct reference images (selected 
using deep feature similarity). This data augmentation procedure further 
boosted the size of our dataset at a relatively low annotation cost. 

\begin{figure*}
\centering
\includegraphics[width=\textwidth]{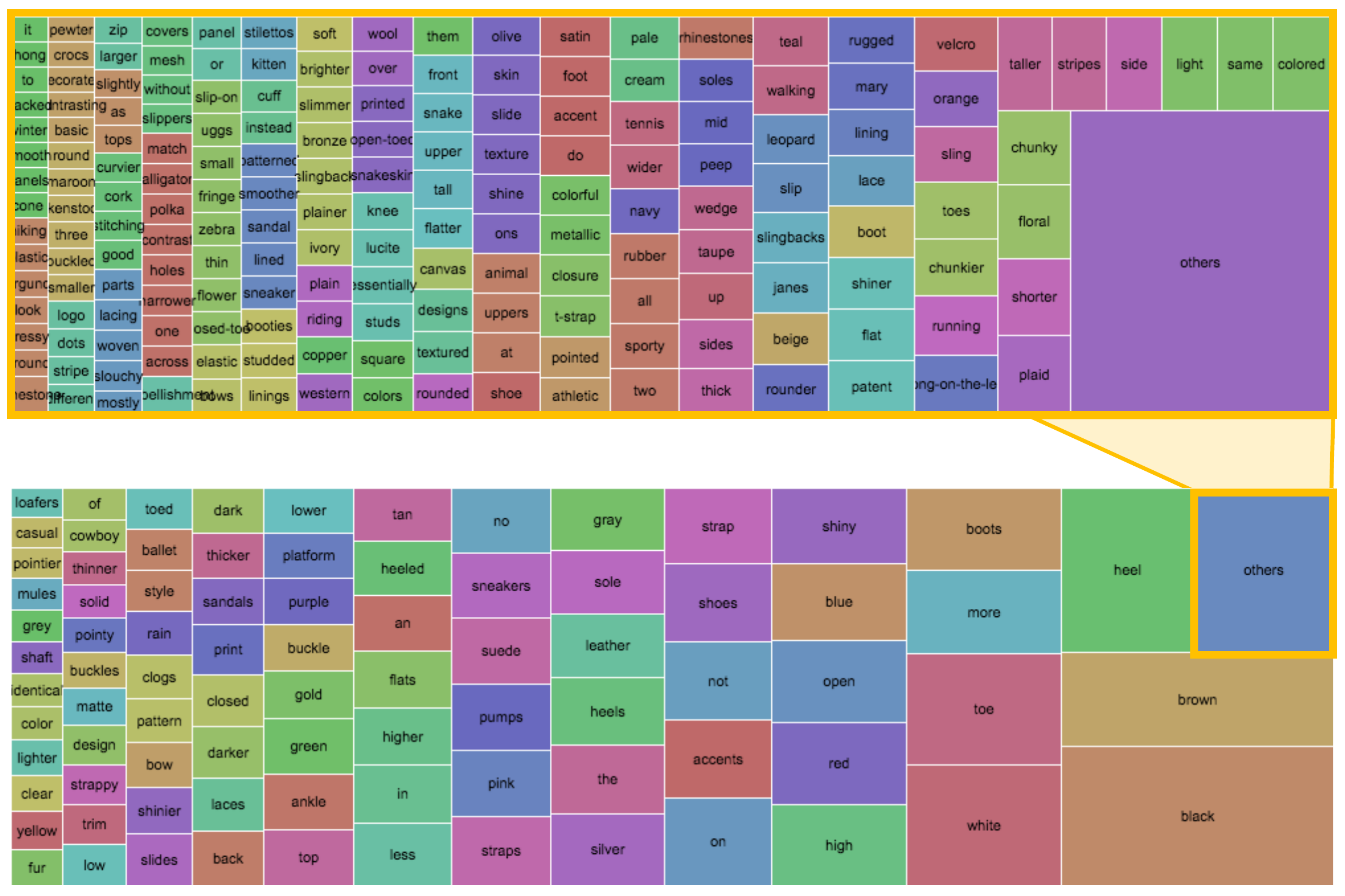}
\caption{Visualization of the rich vocabulary discovered from the 
relative captioning dataset. The size of each rectangle is proportional to the word count of the corresponding
word. }
\label{fig:wordDistr}

\end{figure*}

\section{Dataset analysis}
\label{sec:dataset_analysis}

Figure~\ref{fig:relativeExample}(a) shows the length distribution of the collected
captions. Most captions are very concise (between 4 to 8 words), yet composing
a large body of highly rich vocabularies as shown in Figure~\ref{fig:wordDistr}
\footnote{A few high-frequency words are removed from this chart, 
including "has/have", "is/are", "a", "with".} .
Interestingly, although annotators have 
the freedom to give feedback in terms of comparison on 
a single visual attribute (such as ``\emph{is darker}", 
``\emph{is more formal}"), most feedback expressions consist of 
compositions of multiple phrases that often include spatial or structural details (Table~\ref{tab:phrases}). 

Examples of the collected relative expressions are shown in 
Figure~\ref{fig:relativeExample}(b). We observed that, in some cases, 
users apply a concise phrase to describe the key visual difference (first example); 
but most often, users adopt more complicated phrases (second and third examples). 
The benefit of using free-form feedback can be seen
in the second example: when the two shoes are exactly the same on most attributes 
(white color, flat heeled, clog shoes), the user resorts to using composition of 
a fine-grained visual attribute (``\emph{holes}") with spatial reference (``\emph{on the top}"). Without free-form dialog based feedback, this intricate visual 
difference would be hard to convey.

\begin{table*}
\begin{center}
\small
  \begin{tabular}{p{4cm} | p{4cm} | p{4.4cm}}
    \hline
    {\bf Single Phrase } & \bf{Composition of Phrases } & \bf{Propositional Phrases} \\ 
    {\bf (36\%)} & \bf{(63\%)} & \bf{(40\%)} \\\hline\hline
    
 are brownish & is \textcolor{NavyBlue}{more athletic} and is \textcolor{NavyBlue}{white} & is lower \textcolor{NavyBlue}{on the ankle} and blue \\ \hline
    have a zebra print & has \textcolor{NavyBlue}{a larger sole} and is \textcolor{NavyBlue}{not a high top} & have rhinestones \textcolor{NavyBlue}{across the toe} and a strap \\\hline
    have a thick foot sheath & has \textcolor{NavyBlue}{lower heel} and \textcolor{NavyBlue}{exposes more foot and toe} & are brown \textcolor{NavyBlue}{with a side cut out} \\\hline
    are low-top canvas sneakers& is \textcolor{NavyBlue}{white}, and \textcolor{NavyBlue}{has high heels, not platforms} & is in neutrals \textcolor{NavyBlue}{with buckled strap} and flatter toe \\\hline
    have polka dot linings & is \textcolor{NavyBlue}{alligator}, \textcolor{NavyBlue}{not snake print}, and \textcolor{NavyBlue}{a pointy tip} & is more rugged \textcolor{NavyBlue}{with textured sole} \\
    \hline
  \end{tabular}
\end{center}
 \caption{Examples of relative expressions. Around two thirds of the collected expressions 
contain composite feedback on more than one types of visual feature. And 40\%
of the expressions contain propositional phrases that provide information 
containing spatial or structural details.}
\label{tab:phrases}
\end{table*}

\section{Human Evaluation of Relative Captioning Results}
\label{sec:app_relative}
\begin{figure}
\begin{center}
\includegraphics[width=.6\linewidth]{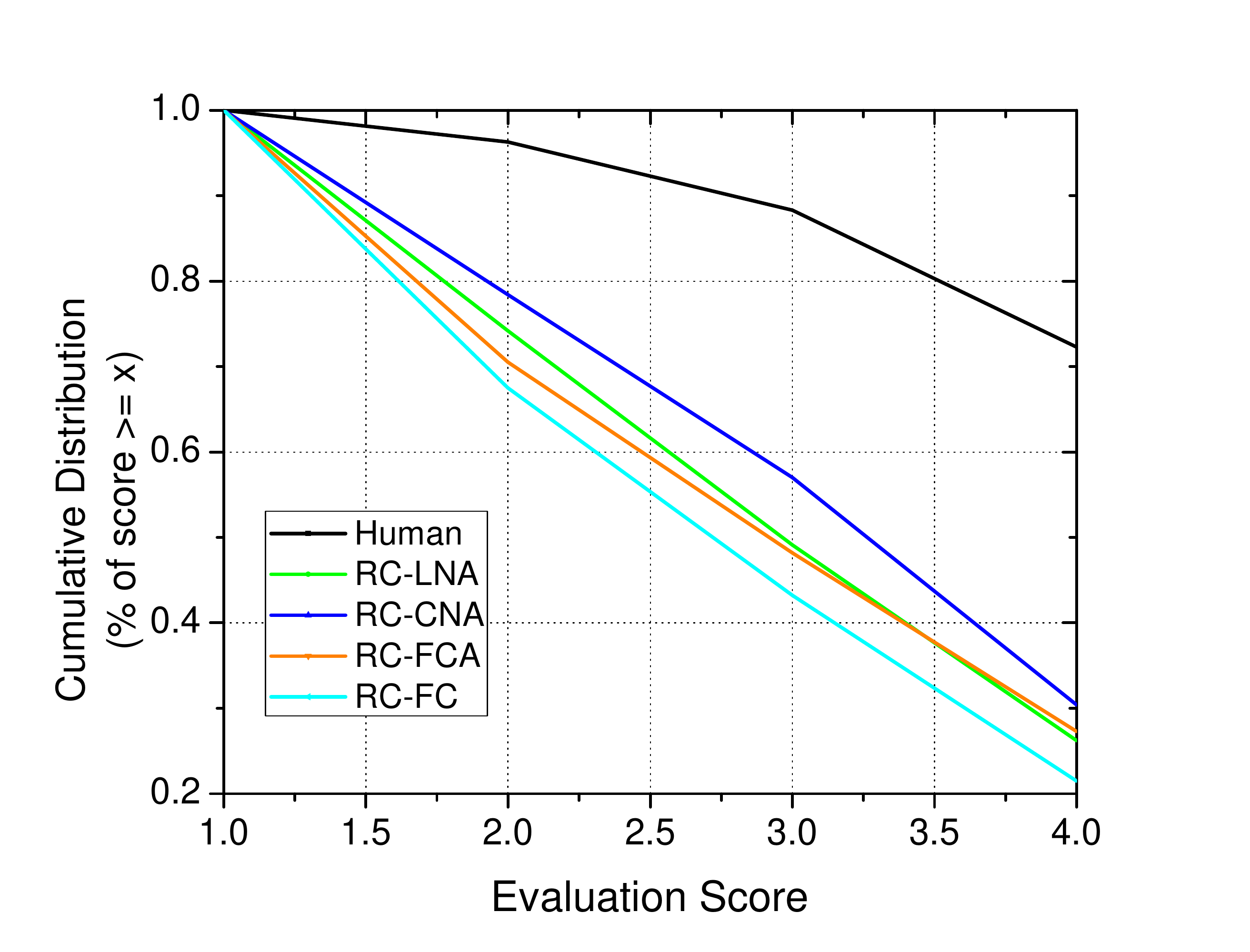}
\caption{Ratings of relative captions provided by 
humans and different relative captioner
models. The raters were asked to give a score from 1 to 4 on the
quality of the captions: no errors (4), minor errors (3), 
somewhat related (2) and unrelated (1). }
\label{fig:human}
\end{center}
\end{figure}

We tested a variety of relative captioning models based on different 
choices of feature fusion and the use of attention mechanism. 
Specifically, we tested one
{\em Show and Tell}~\cite{vinyals2015show} based model,
\textbf{RC-FC} (using concatenated deep features as input), 
and three {\em Show, Attend and Tell}~\cite{icml2015_xuc15} based models,
including \textbf{RC-FCA} (feature concatenation), \textbf{RC-LNA} (feature fusion using 
a linear layer) and \textbf{RC-CNA} (feature fusion using a convolutional
layer). For all methods, we adopted the architecture of ResNet101~\cite{He2015}
pre-trained on ImageNet to extract deep feature representation. 

\begin{figure}
\begin{center}
\includegraphics[width=\linewidth]{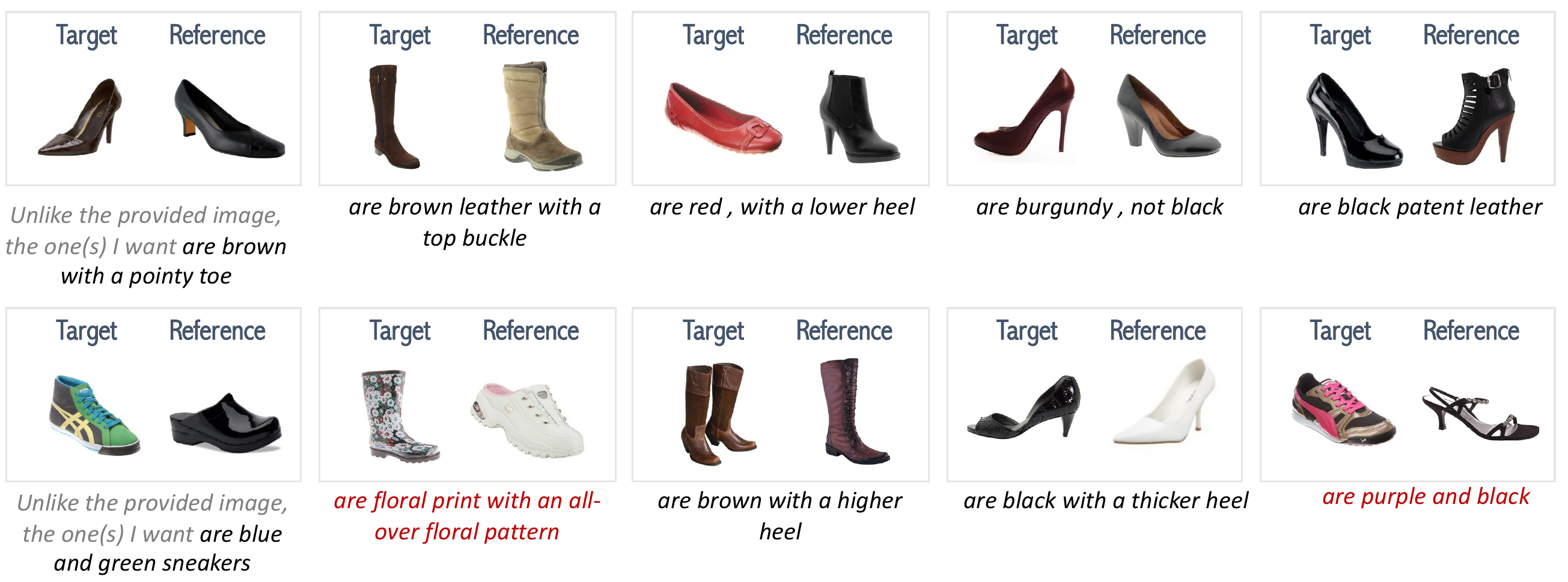}
\caption{Examples of generated relative captions using \textbf{RC-FCA}. Red fonts highlight
inaccurate or redundant descriptions.}
\label{fig:more_captions}
\end{center}
\end{figure}

We report several common quantitative metrics to compare the quality of generated
captions in Table~\ref{tab:scores}. Given the intrinsic flexibility in describing visual differences between two images, and the lack of comprehensive variations 
of human annotations for each pair of images, we found that common image captioning 
metrics does not provide reliable evaluation of the 
actual quality of the generated captions.
Therefore, to better evaluate the caption quality, 
we directly conducted human evaluation, following the same rating scheme used in
\cite{vinyals2015show}. We collected user ratings on relative captions generated by
each model and those provided by humans on $1000$ image pairs. 
Both quantitative results and human evaluation (Figure~\ref{fig:human}) suggest that
all relative captioning models produced similar performance with \textbf{RC-CNA} 
exhibiting marginally better performance. It is also noticeable that there is a gap 
between human provided descriptions and all automatically generated captions,
and we observed some captions with incorrect attribute descriptions or are not
entirely sensible to humans, as shown in Figure~\ref{fig:more_captions}.
This indicates the inherent complexity of task of relative image
captioning and room for improvement of the user simulator,
which will lead to more robust and generalizable dialog agents. 

\begin{table}
  \centering
   \caption{Quantitative metrics of generated relative captions on Shoes dataset.} 
  \begin{tabular}{l|c|c|c}
  \hline
  & BLEU-1 & BLEU-4 & ROUGE  \\
  \hline
  RC-CNA & 32.5 & 11.2 & 45.4 \\
  RC-LNA & 30.7 & 10.7 & 43.2 \\
  RC-FCA & 29.6 & 10.3 & 42.9 \\
  RC-FC & 26.3 & 8.8 & 40.4 \\
  \hline
  \end{tabular}
 \label{tab:scores}
\end{table}

\section{Experimental Configurations}
\label{sec:config}
Since no official training and testing data split was reported on {\em Shoes} 
dataset, we randomly selected $10,000$ images as the training set, 
and the rest $4,658$ images as the held-out testing set. The user simulator adopts
the same training and testing data split as our dialog manager: it
was trained using image pairs sampled from the training set with no overlap 
with the testing images. Since the four models for relative image captioning produced
similar qualitative results in the user study, we selected \textbf{RC-FCA} model as our user simulator since it leads to more efficient training time for the dialog manager 
than the \textbf{RC-CNA} model. 
The baseline method, \textbf{RL-SCST}, uses the same network architecture
and the same supervised pre-training step as our dialog manager 
and also utilizes the user simulator for training. The idea of \textbf{RL-SCST} is to 
use test-time inference reward as the baseline for policy gradient learning 
by encouraging policies performing above 
the baseline while suppressing policies under-performing the baseline. 
Given the trained user simulator, we can easily compute the test-time 
rewards for \textbf{RL-SCST} by greedy decoding rather than stochastically sampling 
the image to return at each dialog turn. 

For all methods, the embedding dimensionality of the feature space is set to $D= 256$;
the MLP layer of the image encoder is finetuned using the single image captions 
to better capture the domain-specific image features. For \textbf{SL} training, 
we used the ADAM optimizer with an initial learning rate of $0.001$ and the 
margin parameter $m$ is set to $0.1$. For all reinforcement 
learning based methods, we employed the RMSprop optimizer with an initial learning 
rate of $10^{-5}$, and the discount factor is set to $1$. 
For our dialog manager, we set the number of nearest neighbors
as $3$ for the Candidate Generator. 
 
\begin{figure*}
\centering
\includegraphics[width=.95\textwidth]{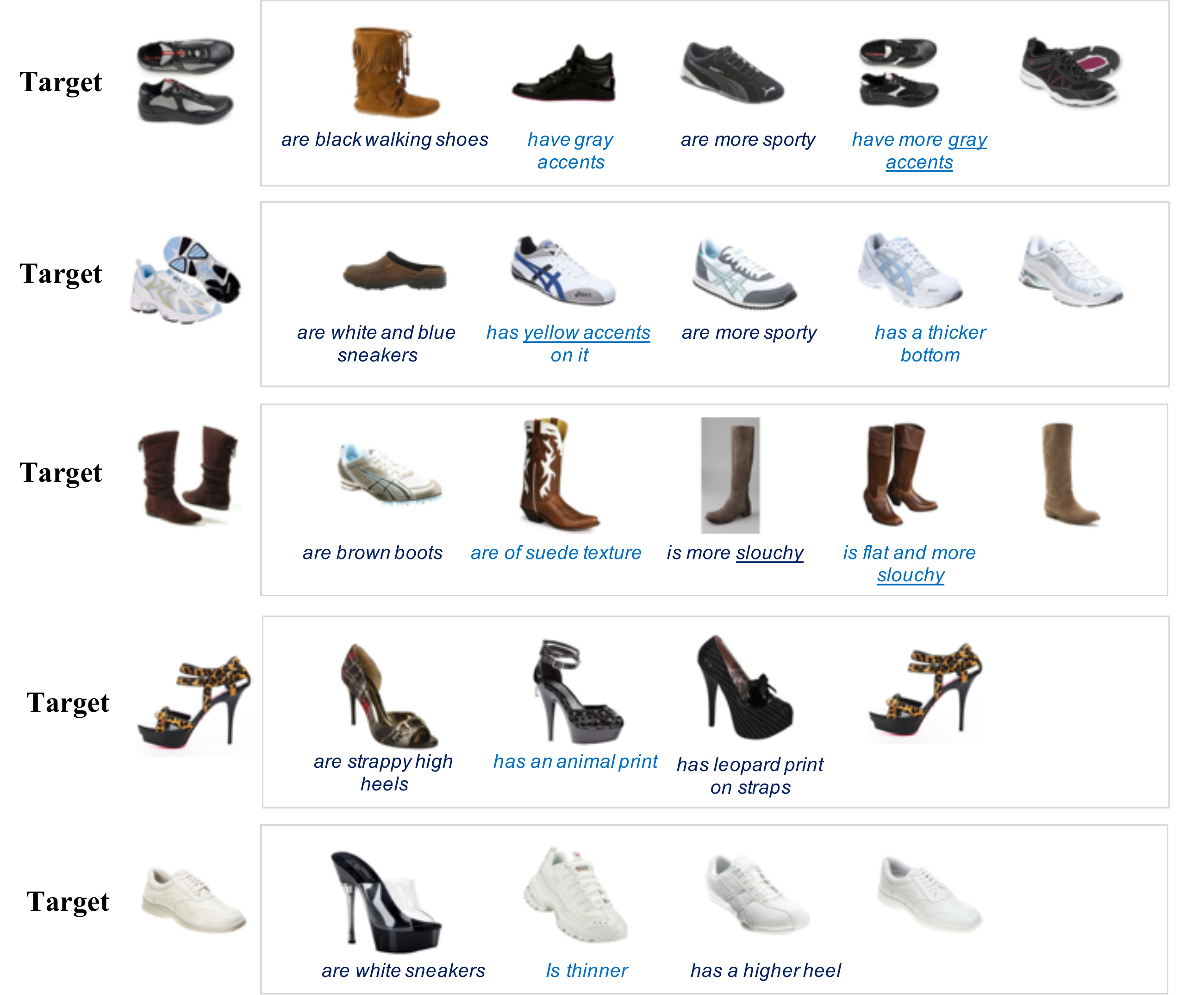}
\caption{Examples of users interacting with the proposed dialog manager system.
User feedbacks are shown below the corresponding images. ``\emph{Unlike the provided image, the ones I want"} is omitted from each sentence for brevity.}
\label{fig:qualitative}
\end{figure*}
\section{Discussions on the Dialog Manager}
\label{sec:qualitative}
In this section, we provide more discussions on the proposed dialog 
manager framework and point out a few directions for improvement. 

\textbf{Dialog-based User Interaction.} 

\begin{wrapfigure}{r}{7cm}
\centering
\includegraphics[width=.4\textwidth]{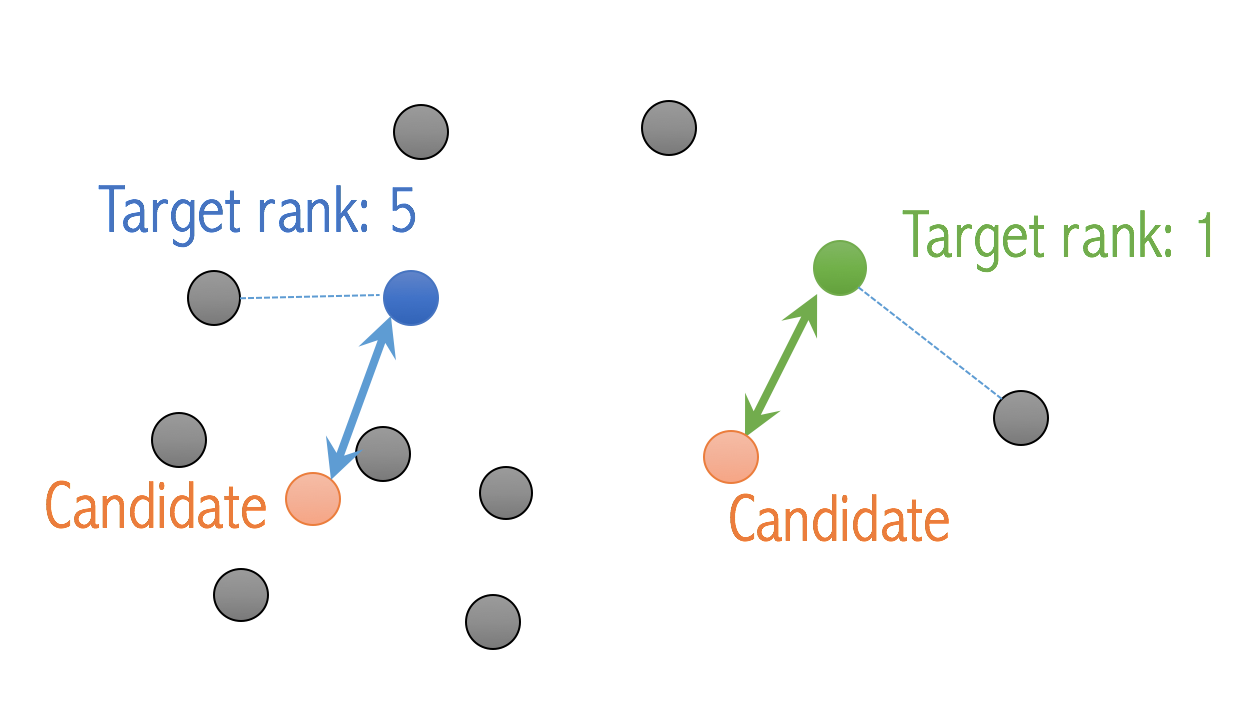}
\caption{Illustration of the triple loss objective and the ranking objective. }
\label{fig:intuition}
\end{wrapfigure}

Figure~\ref{fig:qualitative} shows more examples of the dialog interactions 
on human users. In all examples, the target image reached a final
ranking within the top $100$ images (about $97\%$ in ranking percentile) within
five dialog turns. These examples indicate that, visible 
improvement of retrieval results often comes from a flexible combination of 
direct reference to distinctive visual attributes of the target image, 
and comparison to the candidate image based on relative attributes. 
Ideally, feedback based on a pre-defined attribute set can achieve 
similar performance if the attribute vocabulary is sufficiently 
comprehensive and descriptive (which often consists of hundreds of 
words as in our footwear retrieval application). 
But in practice, it is infeasible to ask the user to scroll through a list of 
hundreds of attribute words and select the optimal one to provide feedback on.

Further, we observe that the system tends to be less responsive to 
certain low-frequency words generated by the use simulator (such as 
``slouchy'' in the third example). This is as expected, since the dialog manager 
is trained on the user simulator, which in itself has limitations
(such as the fixed size of vocabulary after being trained, and the lack of memory for
dialog history). We are interested in finetuning the dialog manager on 
real users, so that it can directly adapt to new vocabularies from the user. 
In summary, results on real users demonstrated that free-form dialog feedback 
is able to capture various types of visual differences with great 
lexical flexibility and can potentially result in valuable applications
in real-world image retrieval systems. 

\textbf{Dialog Manager Learning Framework.} One main advantage of the proposed 
RL based framework is to train the agent end-to-end
with a non-differentiable objective function (the target image rank).
While triplet loss based objective makes it efficient to pre-train the dialog
manager, it still deviates from the ranking objective. 
As illustrated in Figure~\ref{fig:intuition}: two examples exhibit similar triplet loss 
objectives, but the target image ranks differ greatly.

We noticed that the dialog manager based on the current learning architecture 
sometimes forgets information from past 
turns. For example, in the second example of Figure~\ref{fig:qualitative}, the
second turn imposes a ``yellow accents'' requirement to the target image. 
While this feedback is reflected in the immediate next turn, it is missing
from the later turns of the dialog. We think that model architectures which 
better incorporates the dialog history is able to alleviate this issue.
We could in principle investigate more variations of the network 
design to further improve its performance. Overall, the proposed network 
architecture is effective in demonstrating the applicability of dialog-based 
interactive image retrieval.

\end{document}